\documentclass[final]{cvpr}

\usepackage{times}
\usepackage{epsfig}
\usepackage{graphicx}
\usepackage{amsmath}
\usepackage{amssymb}

\usepackage{booktabs}
\usepackage{multirow}
\usepackage{amsthm}
\theoremstyle{definition}
\newtheorem{definition}{Criterion}[section]
\usepackage{algorithmic} 
\usepackage[ruled, vlined]{algorithm2e}

\SetKwInput{KwInput}{Input}               
\SetKwInput{KwOutput}{Output}

\DeclareMathOperator*{\argmax}{argmax}

\usepackage{mathtools}
\DeclarePairedDelimiter\abs{\lvert}{\rvert}%

\usepackage[pagebackref=true,breaklinks=true,colorlinks,bookmarks=false]{hyperref}

\def\cvprPaperID{2175} 
\def\confYear{CVPR 2021}

\begin{document}

\title{Jo-SRC: A Contrastive Approach for Combating Noisy Labels}

\author{Yazhou Yao$^{1}$,
        Zeren Sun$^{1}$,
        Chuanyi Zhang$^{1}$, 
        Fumin Shen$^{2}$,
        Qi Wu$^{3}$, 
        Jian Zhang$^{4}$, 
        Zhenmin Tang$^{1}$\\
$^1$Nanjing University of Science and Technology, Nanjing, China\\
$^2$University of Electronic Science and Technology of China, Chengdu, China\\
$^3$The University of Adelaide, Adelaide, Australia\\ 
$^4$University of Technology Sydney, Sydney, Australia\\
}

\maketitle

\pagestyle{empty}  
\thispagestyle{empty} 

\begin{abstract}
Due to the memorization effect in Deep Neural Networks (DNNs), training with noisy labels usually results in inferior model performance. Existing state-of-the-art methods primarily adopt a sample selection strategy, which selects small-loss samples for subsequent training. However, prior literature tends to perform sample selection within each mini-batch, neglecting the imbalance of noise ratios in different mini-batches. Moreover, valuable knowledge within high-loss samples is wasted. To this end, we propose a noise-robust approach named Jo-SRC (\textbf{Jo}int Sample \textbf{S}election and Model \textbf{R}egularization based on \textbf{C}onsistency). Specifically, we train the network in a contrastive learning manner. Predictions from two different views of each sample are used to estimate its ``likelihood'' of being clean or out-of-distribution. Furthermore, we propose a joint loss to advance the model generalization performance by introducing consistency regularization. Extensive experiments have validated the superiority of our approach over existing state-of-the-art methods. The source code and models have been made available at \url{https://github.com/NUST-Machine-Intelligence-Laboratory/Jo-SRC}.
\end{abstract}

\section{Introduction}

\begin{figure}[t]
\centering
\includegraphics[width=\linewidth]{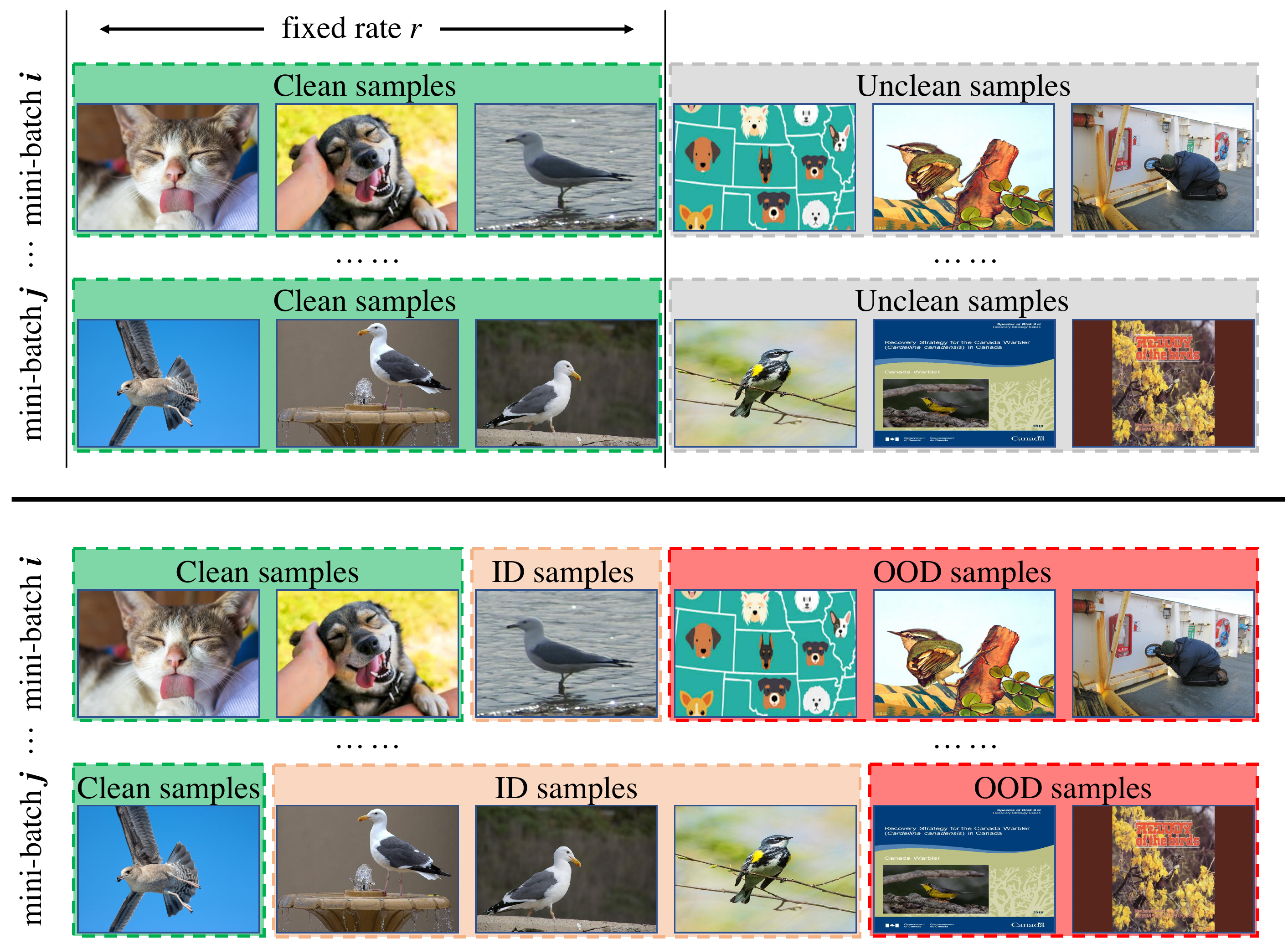}
\caption{Existing small-loss based sample selection methods (\textbf{upper}) tend to regard a human-defined proportion of samples within each mini-batch as clean ones. They ignore the fluctuation of noise ratios in different mini-batches. On the contrary, our proposed method (\textbf{bottom}) selects clean samples in a global manner. Moreover, in-distribution (ID) noisy samples and out-of-distribution (OOD) ones are also selected and leveraged for enhancing the model generalization performance.}
\label{fig:motivation}
\vspace{-0.12cm}
\end{figure}

DNNs have recently lead to tremendous progress in various computer vision tasks \cite{krizhevsky2012imagenet,ren2015faster,yao2021weakly,redmon2017yolo9000,xie2020region,luo2019segeqa}. These successes largely attribute to large-scale datasets with reliable annotations (\eg, ImageNet \cite{deng2009}). However, collecting well-annotated datasets is extremely labor-intensive and time-consuming, especially in domains where expert knowledge is required (\eg, fine-grained categorization \cite{cub200-2011,inat17}). The high cost of acquiring large-scale well-labeled data poses a bottleneck in employing DNNs in real-world scenarios.

As an alternative, employing web images to train DNNs has received increasing attention recently \cite{liu2021exploiting,yang2018recognition,yao2020bridging,tanaka2018joint,yao2018extracting,yao2020exploiting,zhang2020web,zhang2020data,sun2020crssc}. 
Unfortunately, whereas web images are cheaper and easier to obtain via image search engines \cite{fergus2010learning,schroff2010harvesting,yao2017exploiting,yao2016domain}, they usually yield inevitable noisy labels due to the error-prone automatic tagging system or non-expert annotations \cite{niu2018webly,sun2020crssc,yao2018extracting,yao2019towards}.
A recent study has suggested that samples with noisy labels would be unavoidably overfitted by DNNs and consequently cause performance degradation \cite{motivation2017,zhang2016understanding}.

To alleviate this issue, many methods have been proposed for learning with noisy labels.
Early approaches primarily attempt to correct losses during training. 
Some methods correct losses by introducing a noise transition matrix \cite{sukhbaatar2015iclr,patrini2017making,goldberger2017,hendrycks2018using}. However, estimating the noise transition matrix is challenging, requiring either prior knowledge or a subset of well-labeled data. 
Some methods design noise-robust loss functions which correct losses according to predictions of DNNs \cite{reed2015training,zhang2018generalized,tanaka2018joint}. However, these methods are prone to fail when the noise ratio is high.

Another active research direction in mitigating the negative effect of noisy labels is training DNNs with selected or reweighted training samples \cite{mentornet,ren2018learning,decoupling,coteaching,coteachingplus,wei2020combating,sun2020crssc}. 
The challenge is to design a proper criterion for identifying clean samples.
It has been recently observed that DNNs have a memorization effect and tend to learn clean and simple patterns before overfitting noisy labels \cite{motivation2017,zhang2016understanding}. 
Thus, state-of-the-art methods (\eg, Co-teaching \cite{coteachingplus}, Co-teaching+ \cite{coteachingplus}, and JoCoR \cite{wei2020combating}) propose to select a human-defined proportion of small-loss samples as clean ones.
Although promising performance gains have been witnessed by employing the small-loss sample selection strategy, these methods tend to assume that noise ratios are identical among all mini-batches. Hence, they perform sample selection within each mini-batch based on an estimated noise rate. However, this assumption may not hold true in real-world cases, and the noise rate is also challenging to estimate accurately (\eg, Clothing1M \cite{xiao2015learning}).
Furthermore, existing literature mainly focuses on closed-set scenarios, in which only in-distribution (ID) noisy samples are considered.
In open-set cases (\ie, real-world cases), both in-distribution (ID) and out-of-distribution (OOD) noisy samples exist. 
High-loss samples do not necessarily have noisy labels.
In fact, hard samples, ID noisy ones, and OOD noisy ones all produce large loss values, but the former two are potentially beneficial for making DNNs more robust \cite{sun2020crssc}.

Motivated by the self-supervised contrastive learning \cite{chen2020simple,grill2020bootstrap}, we propose a simple yet effective approach named Jo-SRC (\textbf{Jo}int Sample \textbf{S}election and Model \textbf{R}egularization based on \textbf{C}onsistency) to address aforementioned issues.
Specifically, we first feed two different views of an image into a backbone network and predict two corresponding softmax probabilities accordingly. 
Then we divide samples based on two likelihood metrics.
We measure the likelihood of a sample being clean using the Jensen-Shannon divergence between its predicted probability distribution and its label distribution. 
We measure the likelihood of a sample being OOD based on the prediction disagreement between its two views.
Subsequently, clean samples are trained conventionally to fit their given labels.
ID and OOD noisy samples are re-labeled by a mean-teacher model before they are back-propagated for updating network parameters.
Finally, we propose a joint loss, including a classification term and a consistency regularization term, to further advance model performance.
A comparison between Jo-SRC and existing sample selection methods is provided in Figure~\ref{fig:motivation}. The major contributions of this work are:

(1) We propose a simple yet effective contrastive approach named Jo-SRC to alleviate the negative effect of noisy labels. Jo-SRC trains the network with a joint loss, including a cross-entropy term and a consistency term, to obtain higher classification and generalization performance.

(2) Our proposed Jo-SRC selects clean samples globally by adopting the Jensen-Shannon divergence to measure the likelihood of each sample being clean. 
We also propose to distinguish ID noisy samples and OOD noisy ones based on the prediction consistency between samples' different views. ID and OOD noisy samples are relabeled by a mean-teacher network before being used for network update.
	
(3) By providing comprehensive experimental results, we show that Jo-SRC significantly outperforms state-of-the-art methods on both synthetic and real-world noisy datasets. Furthermore, extensive ablation studies are conducted to validate the effectiveness of our approach.

\begin{figure*}[t]
\centering
\includegraphics[width=\linewidth]{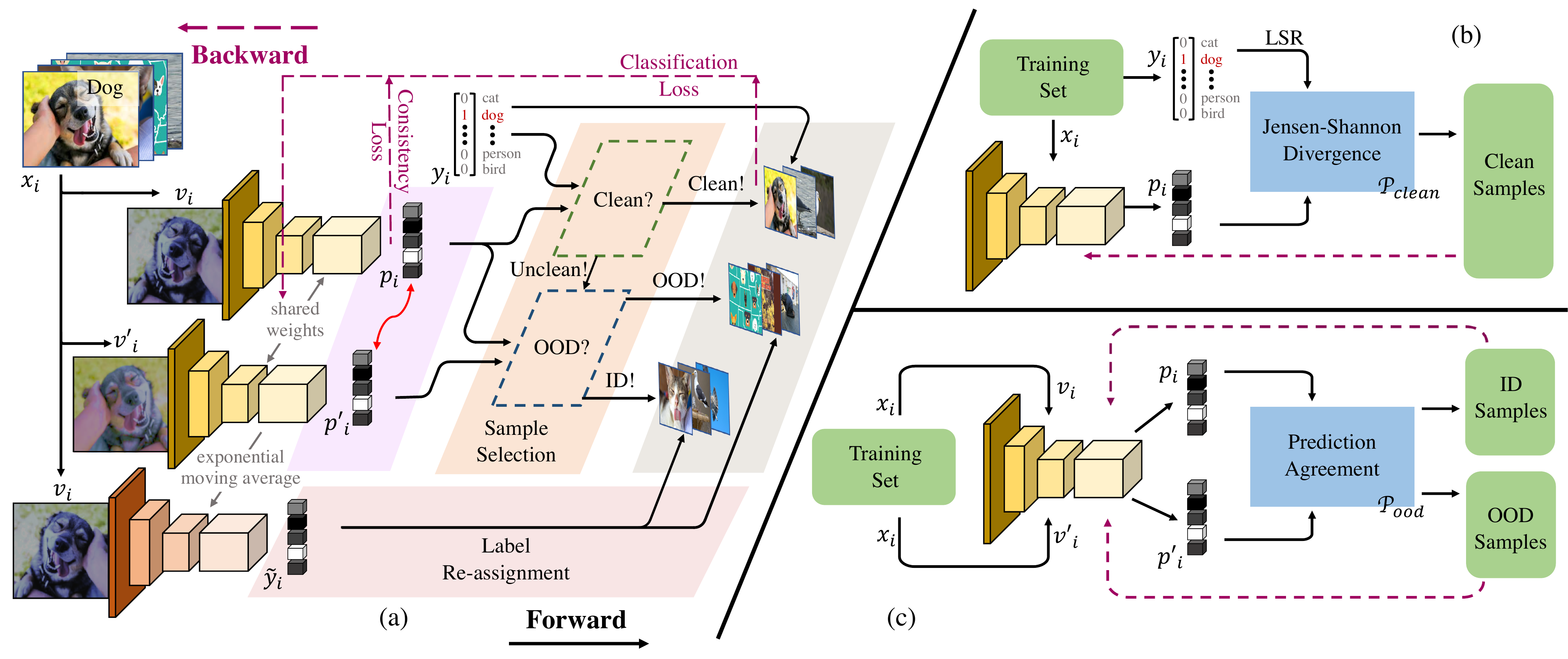}
\caption{The overall framework of our proposed Jo-SRC approach (a), the clean sample selection module (b), and the ID/OOD sample selection module (c).
Each image $x_i$ is augmented into two different views $v_i$ and $v'_i$ before being fed into the backbone network. The network then predicts two probability distributions $\pmb{p_i}$ and $\pmb{p'_i}$ accordingly. 
Afterwards, we obtain the likelihood of $x_i$ being clean $\mathcal{P}_{clean}$ using the Jensen-Shannon (JS) divergence between its predicted distribution $\pmb{p_i}$ and its label distribution $\pmb{y_i}$. 
If $x_i$ is judged as ``unclean'', we obtain its likelihood of being out-of-distribution (OOD) $\mathcal{P}_{ood}$ based on the prediction disagreement between $\pmb{p_i}$ and $\pmb{p'_i}$. 
Finally, $x_i$ is re-labeled as $\pmb{\tilde{y}_i}$ by a mean-teacher model. 
The final objective function is a joint loss, including a classification term and a consistency term.}
\label{fig:pipline}
\vspace{-0.12cm}
\end{figure*}


\section{Related Works}

Existing works on learning with noisy labels can be briefly categorized into the following two subsets \cite{sun2020crssc}: 1) Loss Correction and 2) Sample Selection.

\textbf{Loss correction}. A large proportion of existing literature on training with noisy labels focuses on loss correction approaches. 
Some methods endeavor to estimate the noise transition matrix \cite{sukhbaatar2015iclr,chang2017active,patrini2017making,goldberger2017,hendrycks2018using}. For example, Patrini \etal \cite{patrini2017making} provided a loss correction method to estimate the noise transition matrix by using a deep network trained on the noisy dataset. However, these methods are limited in that the noise transition matrix is challenging to estimate accurately and may not be feasible in real-world scenarios.
Some methods attempt to design noise-tolerant loss functions \cite{reed2015training,zhang2018generalized,tanaka2018joint}. For example, the bootstrapping loss \cite{reed2015training} extended the conventional cross-entropy loss with a perceptual term. However, these methods fail to perform well in real-world cases when the noise ratio is high.

\textbf{Sample Selection}. Another idea of dealing with noisy labels is to select and remove corrupted data. The problem is to find proper sample selection criteria.
It has been shown that DNNs tend to learn simple patterns first before memorizing noisy data \cite{motivation2017,zhang2016understanding}. Resorting to this observation, the small-loss sample selection criterion has been widely adopted: samples with lower loss values are more likely to have clean labels. 
For example, Co-teaching \cite{coteaching} proposed to maintain two networks simultaneously during training, with one network learning from the other network’s selected small-loss samples.
JoCoR \cite{wei2020combating} proposed to use a joint loss, including the conventional cross-entropy loss and the co-regularization loss, to select small-loss samples.
However, above methods select samples within each mini-batch based on a human-defined drop rate. In real-world scenarios, noise ratios in different mini-batches are not guaranteed to be identical, and the drop rate is challenging to estimate. 

\section{The Proposed Method}

\textbf{Background}.
Generally, for a multi-class classification task with $C$ classes, we train DNNs using a labeled dataset $\mathbb{D} = \{(x_i, y_i)| 1 \le i \le N\}$, in which $x_i$ is the $i$-th training sample and $y_i \in \{0, 1\}^C$ is its corresponding one-hot label over $C$ classes. The conventional objective loss function is the cross-entropy between the predicted softmax probability distributions of training samples and their corresponding label distributions:
\begin{equation}\label{eq:celoss}
	\mathcal{L}_{CE} = - \frac{1}{N} \sum_{i=1}^{N} \sum_{c=1}^{C} y_i^c \log(p^c_i),
\end{equation}
in which $ p^c_i $ is a simplified form of $p^c(x_i, \theta)$, denoting the predicted probability of sample $x_i$ for class $c$ given a model with parameters $\theta$.
However, for datasets with noisy labels (\eg, web image datasets), labels are not guaranteed to be correct.
Thus, training DNNs using noisy datasets directly is problematic and usually leads to a dramatic performance drop, given the fact that DNNs have the capability to memorize all training samples, including noisy ones \cite{motivation2017}.

\textbf{Terminology}.
This paper adopts two consistency metrics to reveal how likely each sample could be clean or OOD. We accordingly term them as ``likelihood'', which is different from the concept of ``likelihood'' in statistics.

\subsection{Global clean sample selection}
Regarding samples with small cross-entropy losses as clean ones is one of the most widely-used sample selection criteria. This criterion is justified by the observation, in which DNNs tend to learn clean patterns first and then gradually fit noisy labels \cite{motivation2017,zhang2016understanding}. 
Methods using this criterion (\eg, Co-teaching \cite{coteaching} and Co-teaching+ \cite{coteachingplus}) typically select a pre-defined proportion of small-loss samples within each mini-batch.
Unfortunately, noise ratios in different mini-batches inevitably fluctuate in real-world scenarios. One solution is to record losses for all samples and select samples in the entire training set. However, this becomes impractical when the dataset volume is increasingly huge.

To this end, we propose to reformulate the clean sample selection criterion from another perspective. Specifically, we propose to adopt the Jensen-Shannon (JS) divergence in Eq.~\eqref{eq:jsd_clean} to quantify the difference $d_i$ between the predicted probability distribution $\pmb{p_i} = [p_i^1, p_i^2, ..., p_i^C]$ and the given ground truth label distribution $\pmb{y_i} = [y_i^1, y_i^2, ..., y_i^C]$ of the sample $x_i$ as follows:
\begin{equation}\label{eq:jsd_clean}
	\begin{split}
		d_i & = D_{JS}(\pmb{p_i} \| \pmb{y_i}) \\
 		    & = \frac{1}{2} D_{KL} (\pmb{p_i} \| \frac{\pmb{p_i} + \pmb{y_i}}{2} ) + \frac{1}{2} D_{KL}(\pmb{y_i} \| \frac{\pmb{p_i} + \pmb{y_i}}{2}),   
	\end{split}
\end{equation}
in which $D_{KL}(\cdot \| \cdot)$ is the Kullback-Leibler (KL) divergence function. 
The JS divergence is a measure of differences between two probability distributions. It is known to be bounded in $[0, 1]$, given a base 2 logarithm is used \cite{lin1991divergence}. Therefore, intuitively, we can leverage $d_i$ to measure the ``likelihood'' of $x_i$ being clean as follows:
\begin{equation}\label{eq:prob_clean}
	\mathcal{P}_{clean}(x_i) = 1 - d_i \in [0, 1].
\end{equation}
In fact, $\mathcal{P}_{clean}(x_i)$ reveals the consistency between $\pmb{p_i}$ and $\pmb{y_i}$.
Here, we adopt smoothed label distributions \cite{szegedy2016} in calculating Eq.~\eqref{eq:jsd_clean} to avoid the issue of $\log(0)$.
We finally define our clean sample selection criterion as follows:
\begin{definition}\label{def:clean}
	The sample $x$ is a clean one if its likelihood of being clean $\mathcal{P}_{clean}(x) > \tau_{clean}$.
\end{definition}

\textbf{Why can we select clean samples globally based on $\mathcal{P}_{clean}$?}
Similar to the cross-entropy, the JS divergence is a measurement depicting differences between two probability distributions. 
Since the $\pmb{y_i}$ in Eq.~\eqref{eq:jsd_clean} is not updated in the back-propagation process, the JS divergence between $\pmb{p_i}$ and $\pmb{y_i}$ is equivalent to the cross-entropy between them. Accordingly, our proposed Criterion~\ref{def:clean} is consistent with the small-loss sample selection criterion. 
However, whereas the value of cross-entropy is not constrained, the JS divergence is bounded in $[0, 1]$, making it a natural global selection metric to describe how likely a sample could be clean.
By directly modeling the likelihood of a sample being clean using Eq.~\eqref{eq:prob_clean}, clean samples are selected more efficiently in a global manner, alleviating the issue caused by the imbalance of noise ratios within different mini-batches.

\subsection{Out-of-distribution detection}

Real-world scenarios contain both in-distribution (ID) noisy samples and out-of-distribution (OOD) ones. 
Despite their noisy labels, they can contribute to the model if their labels are re-assigned properly, especially for ID samples.
Therefore, dropping all ``unclean'' samples directly is not data-efficient.

DNNs are usually uncertain about OOD samples when making predictions since their correct labels are outside the task scope. 
Conversely, while ID noisy samples have corrupted labels, they usually lead to consistent model predictions.
Therefore, inspired by the self-supervised contrastive learning \cite{chen2020simple} and agreement maximization principle \cite{sindhwani2005co}, we propose to use the prediction consistency to distinguish OOD and ID samples. 
Specifically, we first generate two augmented views $v_i = T(x_i)$ and $v'_i = T'(x_i)$ from a sample $x_i$ by applying two different image transformations $T(\cdot)$ and $T'(\cdot)$. These two views are subsequently fed into a DNN to produce their corresponding predictions $\pmb{p_i}$ and $\pmb{p'_i}$, respectively. 
Finally, we adopt the consistency between these two predictions to determine if this sample is out-of-distribution or not. 
More explicitly, we define the ``likelihood'' of a sample being out-of-distribution (OOD) as:
\begin{equation}\label{eq:prob_ood}
	\mathcal{P}_{ood}(x_i) = \min (1,  \abs{\argmax_c \pmb{p_i} - \argmax_c \pmb{p'_i}}).
\end{equation}
Consequently, given $\tau_{ood} \in (0, 1)$, our OOD/ID sample selection criterion is defined as follows:

\begin{definition}\label{def:id_ood}
	Given a sample $x$ that is selected as a ``unclean'' one by Criterion~\ref{def:clean}, it is judged as an OOD noisy one if $\mathcal{P}_{ood}(x_i) > \tau_{ood}$ (\ie, its predictions of two differently augmented views disagree). If $\mathcal{P}_{ood}(x_i) \le \tau_{ood}$ (\ie, its predictions of two differently augmented views is consistent), it is deemed as an ID noisy sample.
\end{definition}

\subsection{Label re-assignment}

The proposed Criterion~\ref{def:clean} and \ref{def:id_ood} jointly divide training data into three subsets: a clean subset $\mathbb{S}_{clean}$, an ID subset $\mathbb{S}_{id}$, and an OOD subset $\mathbb{S}_{ood}$. To leverage all training data efficiently, we treat their labels differently before feeding them into the network. 

For samples in $\mathbb{S}_{clean}$, we keep their labels unaltered. To enhance the generalization performance, we adopt the label smoothing regularization (LSR) \cite{szegedy2016} when calculating their losses. Therefore, the label distribution of a clean sample $x_i$ is provided as Eq.~\eqref{eq:label_clean}, given its label $l_i \in \{1, 2, 3, ..., C \}$:
\begin{equation}\label{eq:label_clean}
	\tilde{y}_i^c = \left\{
		\begin{array}{ll}
			1 - \epsilon,			& {c = l_i} \\
			\frac{\epsilon}{C-1},	& {c \neq l_i}	
		\end{array},
	\right.
\end{equation}
in which $\epsilon$ is a hyper-parameter controlling the smoothness of the label distribution. 

For samples in ID subset $\mathbb{S}_{id}$, inspired by the mean-teacher model \cite{tarvainen2017mean}, we use the temporally averaged model (\ie mean-teacher model) to generate reliable pseudo label distributions for providing supervision. Therefore, given an ID sample $x_i$, its pseudo label distribution is provided as:
\begin{equation}\label{eq:label_id}
	\tilde{y}_i^c = p^c(x_i, \theta_{mt}),
\end{equation}
where $\theta_{mt}$ denotes parameters of the mean-teacher model. 

Finally, for samples in $\mathbb{S}_{ood}$, we also use the mean-teacher model to create their corresponding pseudo label distributions. However, since OOD samples' true labels are outside the task scope, the DNN should be highly confused when predicting their label assignments. Therefore, we propose to enforce predictions of OOD samples to fit an approximately uniform distribution for boosting generalization performance. In practice, given an OOD sample $x_i$, we relabel it with the following pseudo label distribution:
\begin{equation}\label{eq:label_ood}
	\tilde{y}_i^c = \frac{e^{p^c(x_i, \theta_{mt})/s}}{\sum_{j=1}^C e^{p^j(x_i, \theta_{mt})/s}},
\end{equation}
in which $s$ is a large scaling constant. In our experiments, we empirically, set $s=10$ to make this label distribution smooth enough (\ie, $ \forall c \in \{1, 2, 3, ..., C\}, \tilde{y}_i^c \approx 1/C $).

It should be noted that the mean-teacher model is not updated via the loss back-propagation. Instead, its parameters  $\theta_{mt}$ is an exponential moving average of $\theta$. Specifically, given a decay rate $\omega \in [0, 1]$, $\theta_{mt}$ is updated in each training step as follows:
\begin{equation}\label{eq:mt_update}
	\theta_{mt} \leftarrow \omega \theta_{mt} + (1- \omega) \theta.
\end{equation}

\begin{algorithm}[t]\small
	\label{alg}
	\SetAlgoLined
	\KwInput{Network $\theta$, mean-teacher $\theta_{mt}$, learning rate $\eta$, iteration $I_{\max}$, epoch $t_w$ and $t_{\max}$.}
	\For{$t = 1, 2, ..., t_{\max}$}
	{
		\For{$\text{iter} = 1, 2, 3, ..., I_{\max}$}
		{
			Sample a mini-batch $\mathbb{B}$ randomly. \\
			Predict $\pmb{p}(x, \theta)$ and $\pmb{p'}(x, \theta)$. \\
			Divide samples into $\mathbb{B}_{clean}$, $\mathbb{B}_{id}$, and $\mathbb{B}_{ood}$ based on Criterion~\ref{def:clean} and \ref{def:id_ood}.\\
			Re-label samples by Eq.~\eqref{eq:label_clean}, \eqref{eq:label_id}, and \eqref{eq:label_ood}.\\
			\eIf {$t_w \le t \le t_{\max}$}{
				Obtain $\mathcal{L}$ using entire $\mathbb{B}$ by Eq.~\eqref{eq:loss_final}.\\
				Update $\theta \leftarrow \theta - \eta \nabla \mathcal{L}$.\\	
			}
			{
				Obtain $\mathcal{L}_{c}$ using only $\mathbb{B}_{clean}$ by Eq.~\eqref{eq:loss_c}.\\
				Update $\theta \leftarrow \theta - \eta \nabla \mathcal{L}_{c} $.\\
			}
			Update $\theta_{mt}$ by Eq.~\eqref{eq:mt_update}.\\
		}
	}
	\KwOutput{Updated network $\theta$.} 
	\caption{Jo-SRC}
\end{algorithm}

\subsection{Consistency regularization}

As stated above, we use each sample's prediction consistency to measure its likelihood of being OOD. 
We follow the intuition that in-distribution samples (including clean ones and noisy ones) tend to produce consistent predictions while out-of-distribution samples do not.
Thus, we propose to use an auxiliary consistency loss as Eq.~\eqref{eq:loss_o} to provide joint supervision 
for enhancing the separability between ID and OOD samples.
\begin{equation}\label{eq:loss_o}
	\mathcal{L}_{o} = \frac{1}{N} \sum_{i=1}^N \rho_i (D_{KL}(\pmb{p_i} \| \pmb{p'_i}) + D_{KL}(\pmb{p'_i} \| \pmb{p_i})),
\end{equation}
in which $\rho_i = 1$ if $x_i \in \mathbb{S}_{clean} \cup \mathbb{S}_{id}$; otherwise, $\rho_i = -1$.

On the one hand, resorting to this additional regularization term, clean samples and ID ones are encouraged to make consistent predictions. 
Meanwhile, this consistency term also enhances the prediction divergence of OOD noisy samples.
Our approach is accordingly able to select clean/ID/OOD samples more effectively.
On the other hand, this auxiliary consistency loss also implicitly promotes representation learning in a self-supervised fashion.

\subsection{The overall framework}

Combining all submodules together, our final objective loss function is 
\begin{equation}\label{eq:loss_final}
	\mathcal{L} = (1 - \alpha) \mathcal{L}_{c} + \alpha \mathcal{L}_{o},
\end{equation}
in which $\alpha$ is a hyper-parameter, and 
\begin{equation}\label{eq:loss_c}
	\mathcal{L}_{c} = \frac{1}{N} \sum_{i=1}^{N} (- \sum_{c=1}^{C} \tilde{y}_i^c \log(p^c_i) - \sum_{c=1}^{C} \tilde{y}_i^c \log(p'^{c}_i)).
\end{equation}
Details of Jo-SRC are shown in Figure~\ref{fig:pipline} and Algorithm~\ref{alg}.

In practice, the model gets increasingly stronger during training and will eventually overfit noisy labels. 
Thus, we proposed to dynamically adjust the selection threshold $\tau_{clean}$ as Eq.~\eqref{eq:dynamic_tauc}:
\begin{equation}\label{eq:dynamic_tauc}
	\tau_{clean} = \left\{
		\begin{array}{ll}
			\frac{t}{t_w}\tau_c,								& {1 \le t \le t_w} \\
			\frac{(t-t_w)\Delta \tau}{t_{\max}-t_w} + \tau_c,	& {t_w < t \le t_{\max}}	
		\end{array},
	\right.
\end{equation}
in which $\Delta \tau = \tau_m - \tau_c$. $\tau_c$ is a hyper-parameter and $\tau_m$ is a large constant ($\tau_m$ is empirically set to 0.95 in our experiments).
Accordingly, more samples will be treated as clean ones in initial epochs so that the model can learn simple and easy patterns from as much samples as possible. As the training proceeds, fewer samples are fed into the model as clean ones for ensuring the quality of learned data.

\begin{table*}[t]
	\centering
	\begin{tabular}{c|c|c|c|c|c|c}
		\toprule
		$\mathfrak{T} - \mathfrak{n}_{c}$		&	Standard	&	Decoupling	&	Co-teaching	&	Co-teaching+	&	JoCoR	&	Jo-SRC	\\
		\midrule
		$\text{Symmetry}	-20\%$		&	35.14 $\pm$ 0.44	&	33.10 $\pm$ 0.12	&	43.73 $\pm$ 0.16	&	49.27 $\pm$ 0.03		&	53.01 $\pm$ 0.04		&	\textbf{58.15} $\pm$ 0.14 \\
		$\text{Symmetry}	-50\%$		&	16.97 $\pm$ 0.40	&	15.25 $\pm$ 0.20	&	34.96 $\pm$ 0.50	&	40.04 $\pm$ 0.70		&	43.49 $\pm$ 0.46		&	\textbf{51.26} $\pm$ 0.11  \\
		$\text{Symmetry}	-80\%$		&	 4.41 $\pm$ 0.14	&	 3.89 $\pm$ 0.16	&	15.15 $\pm$ 0.46	&	13.44 $\pm$ 0.37		&	15.49 $\pm$ 0.98		&	\textbf{23.80} $\pm$ 0.05  \\
		$\text{Asymmetry}-40\%$		&	27.29 $\pm$ 0.25	&	26.11 $\pm$ 0.39	&	28.35 $\pm$ 0.25	&	33.62 $\pm$ 0.39		&	32.70 $\pm$ 0.35		&	\textbf{38.52} $\pm$ 0.20 \\
		\bottomrule
	\end{tabular}
	\caption{Average test accuracy ($\%$) on CIFAR100N-C over the last 10 epochs.}
	\label{table:cifar100nc_comparison}
\end{table*}

\begin{table*}[t]
	\centering
	\begin{tabular}{c|c|c|c|c|c|c}
		\toprule
		$\mathfrak{T} - \mathfrak{n}_{c}$		&	Standard	&	Decoupling	&	Co-teaching	&	Co-teaching+	&	JoCoR	&	Jo-SRC	\\
		\midrule
		$\text{Symmetry}	-20\%$		&	29.37 $\pm$ 0.09	&	43.49 $\pm$ 0.39	&	60.38 $\pm$ 0.22	&	53.97 $\pm$ 0.26		&	59.99 $\pm$ 0.13		&	\textbf{65.83} $\pm$ 0.13 \\
		$\text{Symmetry}	-50\%$		&	13.87 $\pm$ 0.08	&	28.22 $\pm$ 0.19	&	52.42 $\pm$ 0.51	&	46.75 $\pm$ 0.14		&	50.61 $\pm$ 0.12		&	\textbf{58.51} $\pm$ 0.08 \\
		$\text{Symmetry}	-80\%$		&	4.20 $\pm$ 0.07		&	10.01 $\pm$ 0.29	&	16.59 $\pm$ 0.27	&	12.29 $\pm$ 0.09		&	12.85 $\pm$ 0.05		&	\textbf{29.76} $\pm$ 0.09 \\
		$\text{Asymmetry}-40\%$		&	22.25 $\pm$ 0.08	&	33.74 $\pm$ 0.26	&	42.42 $\pm$ 0.30	&	43.01 $\pm$ 0.59		&	39.37 $\pm$ 0.16		&	\textbf{53.03} $\pm$ 0.25 \\
		\bottomrule
	\end{tabular}
	\caption{Average test accuracy ($\%$) on CIFAR80N-O over the last 10 epochs.}
	\label{table:cifar100no_comparison}
\end{table*}

\begin{figure*}[!t]
\centering
\includegraphics[width=0.95\linewidth]{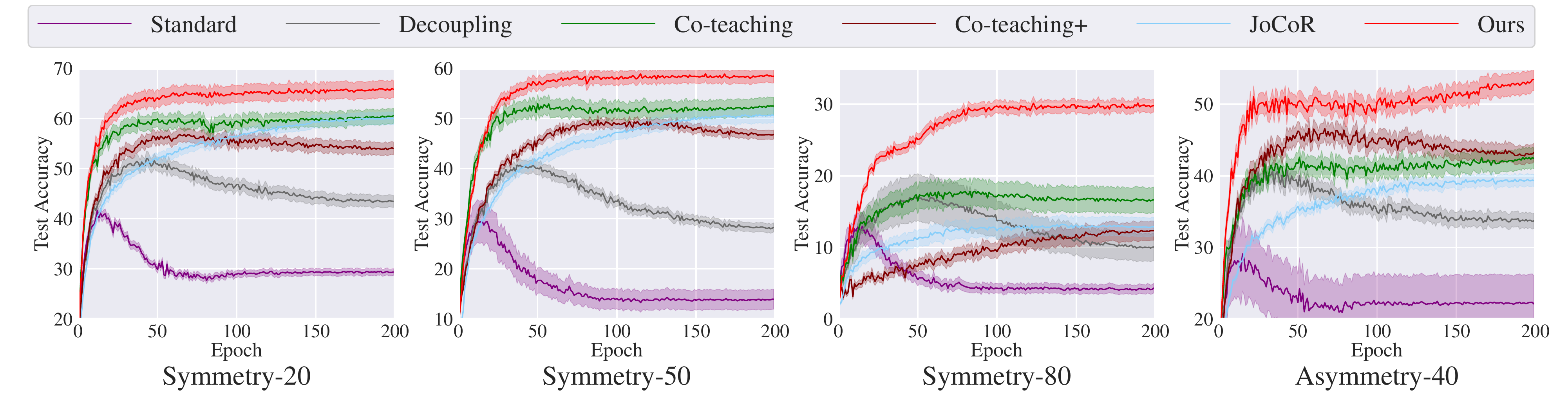}
\caption{Comparison on CIFAR80N-O: test accuracy ($\%$) \textit{vs.} epochs.}
\label{fig:test_acc}
\vspace{-0.12cm}
\end{figure*}

\section{Experiments}

\subsection{Experiment setup}

\textbf{Datasets}. We evaluate Jo-SRC in four benchmark datasets: CIFAR100N-C, CIFAR80N-O, Clothing1M \cite{xiao2015learning}, and Food101N \cite{lee2018cleannet}. 
CIFAR100N-C and CIFAR80N-O are two synthetic datasets created from CIFAR100 \cite{krizhevsky2009}.
Specifically, we follow JoCoR \cite{wei2020combating} to create the closed-set synthetic dataset CIFAR100N-C with a noise ratio $\mathfrak{n}_{c} \in (0, 1)$. The noise type $\mathfrak{T} $ could be either ``Symmetry'' or ``Asymmetry''. 
To create the open-set synthetic dataset CIFAR80N-O, we first regard the last 20 categories in CIFAR100 as out-of-distribution ones. 
Then we create in-distribution noisy samples by randomly corrupting $\mathfrak{n}_{c}$ percentage of remaining samples' labels in a $\mathfrak{T}$ fashion. 
This finally leads to an overall noise ratio $\mathfrak{n}_{all} = 0.2+0.8\mathfrak{n}_{c}$.
Clothing1M and Food101N are two large-scale real-world datasets with noisy labels. 
Details are in supplementary materials. 

\textbf{Evaluation Metrics}.
For evaluating the model classification performance, we take the test accuracy as the evaluation metric.
Besides, we also adopt the label precision as the metric to evaluate our sample selection criteria.

\textbf{Implementation Details}.
Following JoCoR \cite{wei2020combating}, we adopt a 7-layer DNN for CIFAR100N-C and CIFAR80N-O. During training, we use Adam optimizer with a momentum of 0.9. The initial learning rate is 0.001, and the batch size is 128. We train the network for 200 epochs and start to decay the learning rate linearly after 80 epochs. The decay rate in updating the mean-teacher network is set to $\omega = 0.99$. The $\tau_{m}$ and the LSR parameter $\epsilon$ is empirically set to 0.95 and 0.6, respectively.
For Clothing1M, we follow settings in JoCoR \cite{wei2020combating} and use ResNet-18 \cite{he2016deep} with ImageNet pre-trained weights to take a fair comparison with results presented in JoCoR. 
We also conduct experiments using ResNet-50 \cite{he2016deep} and follow experimental settings used in DivideMix \cite{li2020dividemix} for fair comparison.
For Food101N, we use ResNet-50 \cite{he2016deep} pre-trained on ImageNet and follow experimental settings used in DeepSelf \cite{han2019deep}.
All experiments are repeated five times and averaged results are reported accordingly.
Our code implementation is based on PyTorch.

\textbf{Baselines}.
To evaluate Jo-SRC on CIFAR100N-C and CIFAR80N-O, we follow JoCoR \cite{wei2020combating} and compare Jo-SRC with the following state-of-the-art sample selection methods: Decoupling \cite{decoupling}, Co-teaching \cite{coteaching}, Co-teaching+ \cite{coteachingplus}, and JoCoR \cite{wei2020combating}.
To evaluate our approach on Clothing1M, besides the above methods, other state-of-the-art methods like F-correction \cite{patrini2017making}, M-correction \cite{arazo2019unsupervised}, Joint-Optim \cite{tanaka2018joint}, Meta-Cleaner \cite{zhang2019metacleaner}, Meta-Learning \cite{li2019learning}, P-correction \cite{yi2019probabilistic}, and DivideMix \cite{li2020dividemix}	are also compared.
To perform evaluation on Food101N, CleanNet \cite{lee2018cleannet} and DeepSelf \cite{han2019deep} are compared with our approach.
Finally, training directly on noisy datasets is also adopted into comparison as a simple baseline (named as Standard).

\subsection{Comparison on synthetic noisy datasets}

\textbf{Results on CIFAR100N-C}.
Whereas our proposed Jo-SRC method is designed for open-set scenarios, it is also applicable and useful in closed-set cases. 
Comparison in test accuracy with state-of-the-art approaches on CIFAR100N-C is shown in Table~\ref{table:cifar100nc_comparison}. 
For simplicity, the results of existing methods are drawn directly from JoCoR \cite{wei2020combating}, and our method is evaluated using the same experimental settings.
From Table~\ref{table:cifar100nc_comparison}, we can observe that our proposed Jo-SRC method consistently outperforms state-of-the-art methods.
Although performance of all methods drops dramatically in the most inferior case (\ie, Symmetry-$80\%$), our methods still obtain the highest test accuracy.

\textbf{Results on CIFAR80N-O}.
CIFAR80N-O is created to simulate the real-world scenario (\ie, open-set problem). 
We present the comparison in test accuracy with state-of-the-art methods on CIFAR80N-O in Table~\ref{table:cifar100no_comparison}.
We implement all these methods with default parameters. Results in Table~\ref{table:cifar100no_comparison} come from experiments under the same experiment settings. 
From this table, we can observe that our Jo-SRC method performs consistently better than other methods. 
In the simplest case (\ie, Symmetry-$20\%$), while all methods work effectively and robustly (except Standard), our method achieves the best test accuracy.
When the noise scenario becomes harder (\ie, Symmetry-$50\%$, and Asymmetry-$40\%$), model performance inevitably starts to drop, especially Decoupling. However, our method is still effective and outperforms other methods.
Finally, when it goes to the most challenging case (\ie, Symmetry-$80\%$), all approaches fail to combat the massive noisy labels. However, Jo-SRC once again achieves significantly higher performance than other methods, demonstrating the superiority of our method in coping with extremely noisy scenarios.
Figure~\ref{fig:test_acc} shows the test accuracy \vs epochs. From this figure, we can observe that Jo-SRC consistently outperforms other methods by a large margin. Moreover, the superiority in the robustness of our method is demonstrated clearly in these curves.

\begin{table}[t]
	\centering
	\begin{tabular}{c|c|c}
		\toprule
		Method										& 	Backbone	&	Test accuracy	\\
		\midrule
		Stardard										&	ResNet-18	&	67.22	\\
		Decoupling \cite{decoupling}					&	ResNet-18	&	68.48	\\
		Co-teaching \cite{coteaching}				&	ResNet-18	&	69.21	\\
		Co-teaching+	 \cite{coteachingplus}			&	ResNet-18	&	59.32	\\
		JoCoR \cite{wei2020combating}				&	ResNet-18	&	70.30	\\
		\midrule
		Stardard										&	ResNet-50	&	69.21		\\
		F-correction \cite{patrini2017making}		&	ResNet-50	&	69.84		\\
		M-correction \cite{arazo2019unsupervised}	&	ResNet-50	&	71.00		\\
		Joint-Optim \cite{tanaka2018joint}			&	ResNet-50	&	72.16		\\
		Meta-Cleaner \cite{zhang2019metacleaner}		&	ResNet-50	&	72.50		\\
		Meta-Learning \cite{li2019learning}			&	ResNet-50	&	73.47		\\
		P-correction \cite{yi2019probabilistic}		&	ResNet-50	&	73.49		\\
		DivideMix \cite{li2020dividemix}				&	ResNet-50	&	74.76		\\
		\midrule
		Jo-SRC										&	ResNet-18	&	\textbf{71.78}	\\
		Jo-SRC										&	ResNet-50	&	\textbf{75.93}	\\  
		\bottomrule
	\end{tabular}
	\caption{Comparison with state-of-the-art methods in test accuracy ($\%$) on Clothing1M.}
	\label{table:c1m_r18_r50}
\end{table}

\begin{table}[t]
	\centering
	\begin{tabular}{c|c|c}
		\toprule
		Method											&	Backbone	&	Test accuracy	\\
		\midrule
		Stardard										&	ResNet-50	&	84.51			\\
		CleanNet $\omega_{hard}$ \cite{lee2018cleannet}	&	ResNet-50	&	83.47			\\
		CleanNet $\omega_{soft}$ \cite{lee2018cleannet}	&	ResNet-50	&	83.95			\\
		DeepSelf	 \cite{han2019deep}						&	ResNet-50	&	85.11			\\
		\midrule
		Jo-SRC											&	ResNet-50	&	\textbf{86.66}	\\
		\bottomrule
	\end{tabular}
	\caption{Comparison with state-of-the-art methods in test accuracy ($\%$) on Food101N using ResNet-50.}
	\label{table:f101n-r50}
\end{table}

\subsection{Comparison on real-world noisy datasets}
\textbf{Results on Clothing1M}.
To verify the effectiveness of our Jo-SRC, we provide experimental results on real-world scenarios.
Clothing1M is a large-scale real-world dataset. It contains one million training images and yield a $61.54\%$ accuracy of noisy labels \cite{xiao2015learning}.
Table~\ref{table:c1m_r18_r50} shows comparison with state-of-the-art methods using ResNet-18 and ResNet-50 as the backbone network. 
From this table, we can observe that our proposed Jo-SRC approach achieves the best scores on both backbones. 
Using ResNet-18 as the backbone, our method achieves an improvement of $1.48\%$ over the existing state-of-the-art.
When ResNet-50 is adopted, Jo-SRC boosts the test accuracy from $74.76\%$ to $75.93\%$.  

\textbf{Results on Food101N}.
Food101N is another real-world noisy dataset. It contains 310k training images in 101 food categories and also has a large proportion of noisy labels.
Table~\ref{table:f101n-r50} presents the performance comparison with state-of-the-arts.
As shown in Table~\ref{table:f101n-r50}, Jo-SRC achieves the best score and outperforms the state-of-the-art DeepSelf \cite{han2019deep} by $1.55\%$, validating the effectiveness of our approach in dealing with real-world noisy cases.

\begin{figure*}[t]
\centering
\includegraphics[width=0.95\linewidth]{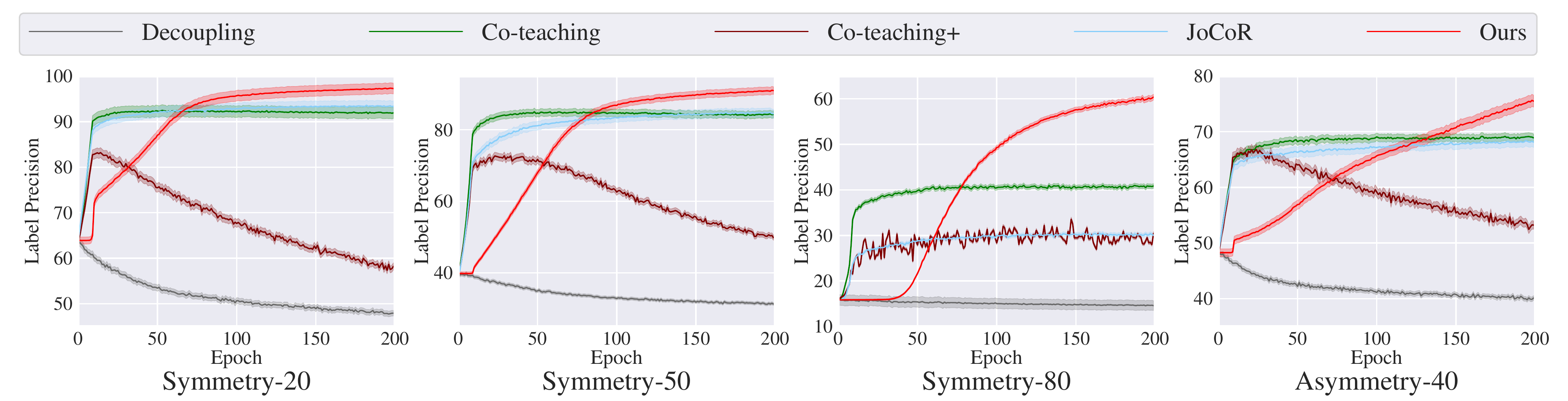}
\caption{Comparison on CIFAR80N-O: precision of clean sample selection ($\%$) \textit{vs.} epochs.}
\label{fig:clean_select}
\vspace{-0.12cm}
\end{figure*}

\begin{figure*}[t]
\centering
\includegraphics[width=0.95\linewidth]{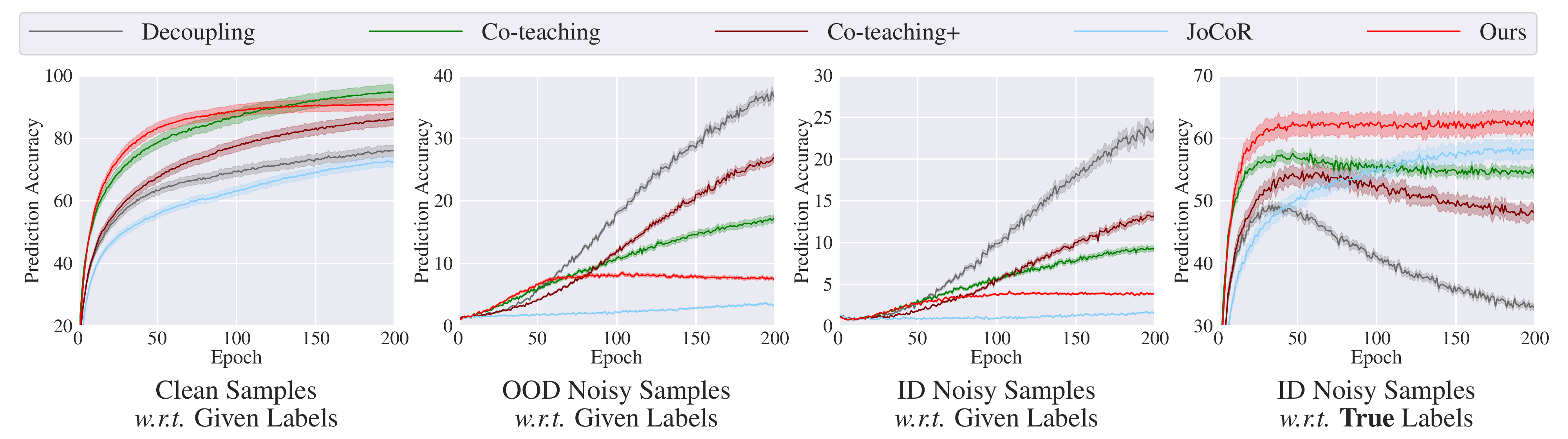}
\caption{The prediction accuracy ($\%$) on different groups of CIFAR80N-O (Symmetry-$20\%$) training data during the training process.}
\label{fig:split_acc}
\vspace{-0.12cm}
\end{figure*}

\begin{table}[t]
	\centering
	\begin{tabular}{ccccc}
		\toprule
		\multirow{2}{*}{Noise}				&	\multicolumn{2}{c}{ID Sample}	&	\multicolumn{2}{c}{OOD Sample}	\\
		\cmidrule(r){2-3}\cmidrule(r){4-5}	&	\textit{best}		&	\textit{last}	& \textit{best}		&	\textit{last}	\\
		\midrule
		$\text{Symmetry}	-20\%$				&	60.91	&	42.62	&	59.54	&	54.38 	\\
		$\text{Symmetry}	-50\%$	 			&	83.86	&	65.92	&	40.70	&	38.75	\\
		$\text{Symmetry}	-80\%$	 			&	96.31	&	72.84	&	26.67	&	24.60	\\
		$\text{Asymmetry}-40\%$	 			&	45.86	&	45.52	&	63.97	&	45.37	\\
		\bottomrule
	\end{tabular}
	\caption{The precision of ID/OOD sample selection on CIFAR80N-O at the \textit{best} and \textit{last} epoch.}
	\label{table:id_ood_select}
\end{table}

\begin{table}[t]
	\centering
	\renewcommand\tabcolsep{15pt}
	\begin{tabular}{l|c}
		\toprule
		Model			&	Test accuracy	\\
		\midrule
		Standard			&	29.37 $\pm$ 0.09			\\
		Jo-SRC-C			&	57.12 $\pm$ 0.33			\\
		Jo-SRC-CI		&	61.32 $\pm$ 0.18			\\
		Jo-SRC-CIO		& 	63.10 $\pm$ 0.07			\\
		Jo-SRC			&	65.83 $\pm$ 0.13		\\
		\bottomrule	
	\end{tabular}
	\caption{Effect of different steps in test accuracy ($\%$) on CIFAR80N-O (Symmetry-$20\%$) over the last 10 epochs.}
	\label{table:module_stacking}
\end{table}

\subsection{Ablation Study}

\textbf{Precision of sample selection}.
The key reason for our approach in obtaining state-of-the-art performance is accurate and reliable sample selection.
To study and verify the superiority of our proposed sample selection strategy, we show the precision of sample selection in Figure~\ref{fig:clean_select} and Table~\ref{table:id_ood_select}.
Figure~\ref{fig:clean_select} presents the precision of clean sample selection \vs epochs. From this figure, Jo-SRC is shown to be highly effective in selecting clean samples accurately and reliably. In all cases, our proposed Jo-SRC achieves the best performance in selecting clean data, compared with state-of-the-art sample selection methods. 
Furthermore, in the most demanding scenario (\ie, Symmetry-$80\%$), while all other methods suffer in finding clean samples, the selection precision of our Jo-SRC increases steadily as the training proceeds. 
These results validate the effectiveness of our clean sample selection strategy. 
Table~\ref{table:id_ood_select} presents the precision in selecting ID/OOD samples. In this table, the \textit{best} and \textit{last} denote the selection precision at the best and last epochs, respectively. Results shown in this table verify the effectiveness of our Jo-SRC in selecting ID/OOD samples.

\textbf{Prediction accuracy of different training samples}.
The memorization effect argues that DNNs would eventually memorize all samples (including noisy ones). 
Therefore, it is critical to prevent networks from overfitting noisy labels when training with noisy datasets.
To further prove the effectiveness of our proposed Jo-SRC, we show the prediction accuracy of different training samples in Figure~\ref{fig:split_acc}. 
As shown in this figure, all methods achieve increasing prediction accuracy on clean samples. JoCoR and our Jo-SRC achieve the lowest prediction accuracy on noisy samples (including both ID ones and OOD ones) \wrt given labels (\ie, noisy labels). This indicates that JoCoR and Jo-SRC perform best in preventing networks from memorizing noisy labels.
Although JoCoR obtains lower prediction accuracy on ID and OOD training samples, it yields an under-fitting issue in clean samples, leading to sub-optimal final test accuracy.
While Co-teaching fits clean samples slightly better than our Jo-SRC, it suffers from overfitting on noisy labels. This causes its final performance decrease in test samples.
Moreover, by observing the last sub-figure, we can find that our Jo-SRC achieves the best prediction accuracy on ID noisy samples \wrt their true labels. This further demonstrates the effectiveness of our sample selection and model regularization, given the fact that ID noisy samples are not supervised by their true labels during training. 

\textbf{Influence of different steps}.
Table~\ref{table:module_stacking} reveals the effect of different steps in our method. 
The Jo-SRC-C denotes the case in which only selected clean samples are adopted in training.
The Jo-SRC-CI denotes the case where clean samples and ID noisy samples are adopted in training.
The Jo-SRC-CIO denotes the case when all samples are adopted in training.
The mean-teacher-based re-labeling is performed accordingly when noisy samples are leveraged in training.
Lastly, the Jo-SRC denotes the final proposed method.
From this table, we can observe that the proposed clean sample selection plays the most crucial role in addressing the label noise issue.
Moreover, appropriately treated noisy samples (including ID and OOD ones) can contribute to the model generalization performance.
Finally, the consistency loss promotes model performance by further regularization.

\section{Conclusion}

In this paper, we proposed a simple yet effective approach named Jo-SRC to address the performance degradation caused by noisy labels.
Jo-SRC trained DNNs in a contrastive manner. 
Clean samples were identified globally based on JS divergence, while ID and OOD noisy samples were distinguished based on consistency.
Samples were selected and divided accordingly for subsequent network learning.
Finally, a joint loss, including a classification term and a consistency regularization term, was proposed to further advance the performance and robustness.
Comprehensive experiments on both synthetic and real-world noisy datasets validated the superiority of the proposed method.

\section*{Acknowledgments}

This work was supported by the National Natural Science Foundation of China (No. 61976116) and Fundamental Research Funds for the Central Universities (No. 30920021135).

{\small
\bibliographystyle{ieee_fullname}
\bibliography{egbib}
}

\end{document}